\documentclass{article}

\usepackage[dblblindworkshop, final]{neurips_2025}
\workshoptitle{Structured Probabilistic Inference and Generative Modeling}

\usepackage[utf8]{inputenc} 
\usepackage[T1]{fontenc}    
\usepackage{hyperref}       
\usepackage{url}            
\usepackage{booktabs}       
\usepackage{amsfonts}       
\usepackage{nicefrac}       
\usepackage{microtype}      
\usepackage{xcolor}         
\usepackage{graphicx}
\usepackage{subcaption}
\usepackage{amsmath}
\usepackage{cleveref}

\usepackage{amsmath,amsfonts,bm,amsthm}

\def\eqref#1{equation~\ref{#1}}

\def\1{\bm{1}}

\def\eps{{\epsilon}}

\def\rvepsilon{{\mathbf{\epsilon}}}
\def\rveps{{\rvepsilon}}

\def\rvx{{\mathbf{x}}}

\def\vx{{\bm{x}}}

\DeclareMathAlphabet{\mathsfit}{\encodingdefault}{\sfdefault}{m}{sl}
\SetMathAlphabet{\mathsfit}{bold}{\encodingdefault}{\sfdefault}{bx}{n}

\newcommand{\bx}{{\bf x}}

\newcommand{\noisyx}{\tilde{\bx}}
\newcommand{\beps}{{\bf{\epsilon}}}
\newcommand{\shorteq}{\mathrel{\mkern-2mu=\mkern-2mu}}
\newcommand{\seq}{\mathrel{\scalebox{0.7}[1]{$\shorteq$}}}
\newcommand{\esm}{\mathcal{L}_\text{ESM}}
\newcommand{\ism}{\mathcal{L}_\text{ISM}}
\newcommand{\dsm}{\mathcal{L}_\text{DSM}}
\newcommand{\esmglobal}{\mathbb{E}_{\noisyx} \left[ \esm(\noisyx,\, \sigma,\, \theta) \right]}
\newcommand{\ismglobal}{\mathbb{E}_{\noisyx} \left[ \ism(\noisyx,\, \sigma,\, \theta) \right]}

\newcommand{\dsmglobal}{\mathbb{E}_\rvx\left[\mathcal{L}_\text{DSM}(\rvx,\, \sigma,\, \theta)\right]}

\newtheorem{theorem}{Theorem}[section]  

\newtheorem{corollary}[theorem]{Corollary}

\newtheorem*{theorem*}{Theorem}
\newtheorem*{lemma*}{Lemma}

\theoremstyle{definition}
\newtheorem{definition}[theorem]{Definition}

\theoremstyle{remark}
\newtheorem{remark}[theorem]{Remark}

\title{A Connection Between Score Matching\\and Local Intrinsic Dimension}

\author{
  Eric Yeats \\
  PNNL \\
  \And
  Aaron Jacobson$^\dagger$ \\
  UNC Chapel Hill \\
  \And
  Darryl Hannan \\
  PNNL \\
  \And
  Yiran Jia$^\dagger$ \\
  UC San Diego \\
  \AND
  Timothy Doster \\
  PNNL \\
  \And
  Henry Kvinge \\
  PNNL \\
  \texttt{\{first\}.\{last\}@pnnl.gov} \\
  \And
  Scott Mahan \\
  PNNL \\
}

\begin{document}

\maketitle

\renewcommand{\thefootnote}{\fnsymbol{footnote}}
\footnotetext[2]{Work done during an internship at Pacific Northwest National Laboratory (PNNL).}

\begin{abstract}
    The local intrinsic dimension (LID) of data is a fundamental quantity in signal processing and learning theory, but quantifying the LID of high-dimensional, complex data has been a historically challenging task. Recent works have discovered that diffusion models capture the LID of data through the spectra of their score estimates and through the rate of change of their density estimates under various noise perturbations. While these methods can accurately quantify LID, they require either many forward passes of the diffusion model or use of gradient computation, limiting their applicability in compute- and memory-constrained scenarios.
    
    We show that the LID is a lower bound on the denoising score matching loss, motivating use of the denoising score matching loss as a LID estimator. Moreover, we show that the equivalent implicit score matching loss also approximates LID via the normal dimension and is closely related to a recent LID estimator, FLIPD. Our experiments on a manifold benchmark and with Stable Diffusion 3.5 indicate that the denoising score matching loss is a highly competitive and scalable LID estimator, achieving superior accuracy and memory footprint under increasing problem size and quantization level.
\end{abstract}

\section{Introduction}

Observations of high-dimensional data which are generated by physics or other natural phenomena tend to inherit low-dimensional structure. This is commonly referred to as the manifold hypothesis, and it underpins central assumptions in machine learning \cite{bengio2013representation}. The fundamental quantity encapsulating the lower dimensional structure of data is the local intrinsic dimension (LID). For a point $x$ on a data manifold, the LID is the local number of dimensions required to losslessly encode the data around $x$.

The LID has clear implications in signal processing, as it determines bounds on how (locally) compressible a distribution is \cite{cover1999elements}. Moreover, the LID is vital to deep learning \cite{lecun2015deep}, where learning from high-dimensional data is made possible by its relatively low-dimensional structure. More specifically, lower LID improves the statistical efficiency of learning - lower dimensional structure makes learning and generalization easier \cite{pope2021intrinsic}. The LID is also a practical tool which has been leveraged in engineering for anomaly detection \cite{yin2024characterizing}, clustering, and segmentation \cite{carter2009local}.

Historically, non-parametric methods have estimated LID by modeling nearby samples with statistical processes \cite{levina2004maximum}, gleaning nearest neighbor information \cite{facco2017estimating}, measuring fractal dimension \cite{grassberger1983measuring}, and calculating simplex skewness \cite{johnsson2014low}. While these methods are effective in simple, small-scale scenarios, they require large amounts of sampled data, are strongly affected by hyperparameter choice, and fail to generalize in low-data settings \cite{tempczyk2022lidl,stanczuk2024diffusion,kamkari2024geometric,yeats2023adversarial}.

Recent works have estimated LID using parametric deep generative models -- they inherit the advantages of deep learning: applicability to big problems and generalization to unseen data. Harnessing the power of deep generative models has led to unprecedented LID estimation capabilities on complex synthetic manifolds and applicability to high-dimensional, real-world problems \cite{tempczyk2022lidl,stanczuk2024diffusion,kamkari2024geometric}. While these methods are more accurate compared with existing non-parametric approaches, they are computationally expensive. This expense comes from requiring multiple models \cite{tempczyk2022lidl}, exhaustive samples \cite{stanczuk2024diffusion}, or gradient computation \cite{kamkari2024geometric} to estimate LID.

Our work introduces an accurate, inexpensive LID estimator based on the denoising score matching (DSM) loss and provides the following contributions:
\begin{itemize}
    \item We prove that the denoising score matching loss is lower bounded by the LID, motivating its use as a scalable LID estimator that does not require exhaustive samples or gradient.
    \item We demonstrate a close relationship between the score matching losses and the current leading estimators, FLIPD \cite{kamkari2024geometric} and the normal bundle method \cite{stanczuk2024diffusion}. We prove that expected FLIPD is also lower bounded by the LID through its connection to the \textit{implicit} score matching loss.
    \item We provide experiments on a manifold benchmark and with Stable Diffusion 3.5 and Stable Diffusion 2 (appendix) which show that the denoising score matching loss is a highly competitive LID estimator. Moreover, it exhibits superior scalability in terms of memory footprint and consistency under model quantization.
\end{itemize}

\section{Background and Related Work}

\paragraph{Denoising Generative Models} Diffusion and flow-based models have achieved state-of-the-art generative modeling capabilities by learning to denoise data \cite{ho2020denoising,esser2024scaling}. The connection between denoising and generative modeling was proven by Vincent \cite{vincent2011connection}, who showed that the denoising objective is a scalable method to learn the \textit{score} of the probability density ($\nabla \log p(\noisyx)$) of noised samples $\noisyx \leftarrow {\bf{x}} + \sigma\beps$, where ${\bf{x}} \sim p({\bf x})$, $\sigma \in \mathbb{R}^+$, and $\beps$ is drawn from a standard Gaussian distribution. For a score function $s_\theta: \mathbb{R}^n \rightarrow \mathbb{R}^n$ parameterized by $\theta$, the denoising objective is equivalent (up to a constant) to the explicit score matching and implicit score matching objectives \cite{hyvarinen2005estimation} for the noised distribution:
\begin{equation}\label{eqn:score_matching}
    \esmglobal = \ismglobal + C_\text{ISM} = \mathbb{E}_{\bf x} \left[ \dsm({\bf x}, \sigma, \theta) \right] + C_\text{DSM},
\end{equation}
where each loss $\mathcal{L}(\vx,\,\sigma,\, \theta)$ is the pointwise version of the loss evaluated at the point $\vx$, and $C_\text{ISM}$ and $C_\text{DSM}$ are constants which do not depend on the parameters $\theta$. The denoising objective $\dsm$ is particularly appealing for training deep generative models, as it does not require oracular knowledge of the true score (required in $\esm$) or expensive computation of $\nabla_{\vx} \cdot s_\theta(\vx)$ (required in $\ism$). Once the score is approximated for many distributions bridging the data distribution and an easy to sample distribution (typically Gaussian), one may employ reverse diffusion samplers \cite{song2020score}, ODE solvers \cite{lipman2022flow}, or Langevin dynamics \cite{song2019generative} to draw samples from the data distribution.

\paragraph{Non-Parametric LID Estimation} Non-parametric local intrinsic dimension estimation methods provide data-driven approaches to estimate the structural dimensionality of high-dimensional datasets without assuming specific parametric forms. The MLE method by Levina and Bickel \cite{levina2004maximum} models the distribution of distances from each point to its k nearest neighbors, estimating intrinsic dimension through maximum likelihood fitting of Poisson distribution parameters to ratios of consecutive nearest neighbor distances. Similarly exploiting nearest neighbor statistics, the TwoNN (Two Nearest Neighbors) \cite{facco2017estimating} method analyzes the ratio of distances to the second and first nearest neighbors, using the empirical distribution of these ratios to infer local dimensionality. Taking a more geometric approach, Expected Simplex Skewness (ESS) \cite{johnsson2014low} estimates dimension by measuring the skewness of volumes of simplices formed by points and their nearest neighbors. Beyond these smooth manifold approaches, fractal dimension methods such as box-counting and correlation dimension estimators \cite{grassberger1983measuring} handle datasets with self-similar or fractal structure.

\paragraph{Parametric LID Estimation} Parametric estimators of local intrinsic dimension typically leverage deep generative models such as diffusion models, flow-matching models, or normalizing flows. LIDL \cite{tempczyk2022lidl}, originally developed with normalizing flows, uses an ensemble of generative models trained on different levels of Gaussian noise. The generative models provide a set of density estimates at a point $\vx$ from which the LID can be retrieved by measuring the rate of change of density estimates under increasing noise perturbations. The normal bundle (NB) method \cite{stanczuk2024diffusion} estimates LID through the connection between the score and the vector to a manifold. The NB method adds $m$ scaled Gaussian noise instances to a point $\vx \in \mathbb{R}^n$ to yield $\noisyx$ and computes the score estimate $s_\theta(\noisyx)$ for each, yielding an ($m,\,n$) matrix of score estimates. The count of non-negligible singular values is taken as the normal dimension, and that can be subtracted from the ambient dimension to yield the LID. The authors recommend using a large number of samples to ensure that there are at least as many samples as the normal dimension. Recently, Kamkari et al.~\cite{kamkari2024geometric} proposed FLIPD, which uses the Fokker-Planck equation from diffusion models to accurately estimate LID. While FLIPD shares LIDL's approach of measuring density change rates under increasing Gaussian noise, it achieves greater efficiency by requiring only a single diffusion model and one Fokker-Planck equation evaluation.

\begin{figure*}[t]
    \begin{subfigure}[t]{0.49\textwidth}
        \centering
        \includegraphics[width=0.95\textwidth]{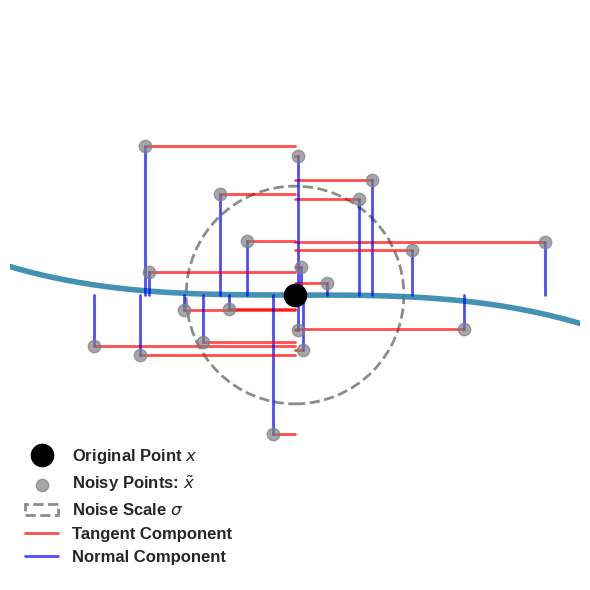}
        \caption{Conceptual depiction of the denoising loss as a LID estimator on a uniformly sampled 1-dimensional manifold. Noise components corresponding to the \textcolor{red}{tangent} space yield an expected squared error of approximately $1$ each, whereas noise components corresponding to the \textcolor{blue}{normal} space yield an expected squared error of approximately $0$ each. Adding up the expected squared error for each dimension yields the LID.}\label{fig:dl_concept}
    \end{subfigure}
    \hfill
    \begin{subfigure}[t]{0.49\textwidth}
        \centering
        \includegraphics[width=0.95\textwidth]{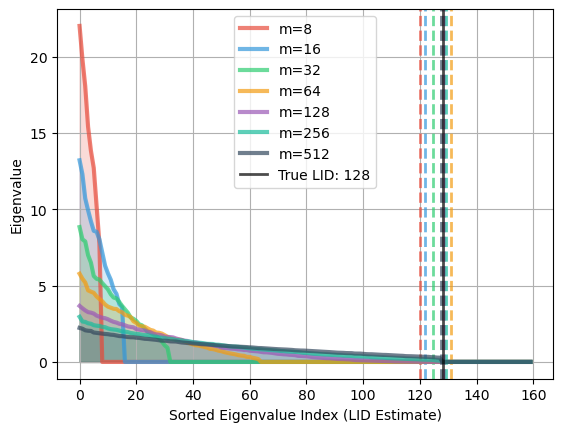}
        \caption{Conceptual link between the error bundle (EB) method (spectra depicted as solid lines) versus the denoising loss method (area under the spectrum). Starting from the the error Gram matrix, the EB method estimates the LID by counting non-negligible eigenvalues, whereas the denoising loss method sums the eigenvalues (area values shown with dashed lines). The denoising loss method is accurate at small sample sizes.}\label{fig:eb_vs_dl}
    \end{subfigure}
    \caption{Conceptual depiction of the denoising loss as a LID estimator (left) and a conceptual link between the denoising loss and the error bundle (EB) method.}
\end{figure*}

\section{Score Matching as a LID Estimator}

We consider training a score model on a single noise level $\sigma \in \mathbb{R}^+$, and we draw connections between the denoising score matching loss and the LID of the data manifold. Let $\rvx$ be a random variable drawn from a $d$-dimensional data manifold $\mathcal{M}$ embedded within $\mathbb{R}^n$. Let $s_\theta: \mathbb{R}^n \rightarrow \mathbb{R}^n$ be the score function parameterized by $\theta$. Furthermore, let $\rveps \sim \mathcal{N}({\bf{0}};\, {\bf{I}})$ be a standard Gaussian random variable in $\mathbb{R}^n$. Recall that the scaled denoising score matching loss \cite{vincent2011connection} is:
\begin{equation}\label{eqn:denoising_loss}
     \dsmglobal := \mathbb{E}_{\rvx}\left[\mathbb{E}_{\rveps \sim \mathcal{N}({\bf{0}};\, {\bf{I}})} \, \sigma^2 \left\| \frac{\rveps}{\sigma} + s_\theta(\rvx + \sigma \rveps) \right\|^2 \right]
     \;= \mathbb{E}_{\rvx,\rveps} \left\| \rveps - \rveps_\theta(\rvx + \sigma \rveps) \right\|^2,
\end{equation}
where $\eps_\theta(\vx)=-\sigma\, s_\theta(\vx)$, the noise prediction parameterization of the score.

\begin{definition}[High-Resolution Assumption]
For a data point $x$ on the manifold $\mathcal{M}$, let $\kappa$ be the maximum principal curvature at $x$, and let $L_p \approx \|\nabla \log p(x)\|^{-1}$ represent the characteristic length scale of density variations on the manifold. We assume the added noise scale $\sigma$ is very small relative to both the geometry and the statistical distribution of the data, such that:
\begin{enumerate}
    \item $\sigma \ll 1/\kappa$ (Negligible Curvature)
    \item $\sigma \ll L_p$ (Locally Constant Density)
\end{enumerate}
\end{definition}
The \textit{negligible curvature assumption} enables local decomposition of the manifold into normal and tangent subspaces, and it is central to our proofs. The \textit{locally constant density} assumption prevents emergence of terms independent of dimension from appearing in our bounds.
\begin{theorem}[DSM Loss Lower Bound]\label{thm:dsm_bound}
Let $x$ be drawn from a $d$-dimensional manifold $\mathcal{M}$ embedded in $\mathbb{R}^n$ with data thickness $\hat{\sigma}$. If the added noise level $\sigma$ satisfies the High-Resolution Assumption ($\sigma \ll 1/\kappa$ and $\sigma \ll \|\nabla \log p(x)\|^{-1}$), then the expected denoising score matching loss is lower bounded by (proof in \Cref{sec:dsm_bound_proof}):
\begin{equation}
    \mathbb{E}_{x}[\mathcal{L}_{DSM}(x, \sigma, \theta)] \ge d + (n-d) \left( \frac{\hat{\sigma}^2}{\sigma^2 + \hat{\sigma}^2} \right).
\end{equation}
\end{theorem}
Here, manifold thickness refers to Gaussian structure of scale $\hat{\sigma}$ within the normal subspace. In plain language, the expected denoising score matching loss is lower bounded by the intrinsic dimension of the manifold plus a ``manifold thickness'' term. The non-negative thickness term inflates dimension estimates in the normal space dimensions.
\begin{corollary}[Tightness at Optimality]\label{cor:tightness}
Assume the data manifold has zero thickness ($\hat{\sigma} = 0$) and the score model is expressive enough to capture the true score. Then, the lower bound is tight. Consequently, the scaled denoising loss of the optimal score model ($\theta=\theta^*$) at any specific point $x \in \mathcal{M}$ is exactly the intrinsic dimension (proof in \Cref{sec:tightness_proof}):
\begin{equation}
    \mathcal{L}_{DSM}(x, \sigma, \theta^*) = d.
\end{equation}
\end{corollary}

In other words, the bound is tight for the class of score models which are expressive enough to capture the true score, and simply computing the denoising loss with an optimally trained score model at an input point $\vx\in\mathcal{M}$ yields the dimension $d$. In practice, the denoising score matching loss is a biased estimator of $d$ since real models have finite capacity and \Cref{thm:dsm_bound} is a lower bound.

\begin{remark}[\textbf{Stratified Manifolds and Local Intrinsic Dimension (LID)}]
    \Cref{thm:dsm_bound} establishes that the population-average loss is lower bounded by the global intrinsic dimension $d$. However, real-world data often lies on a stratified manifold composed of disjoint submanifolds $\mathcal{M}_i$, each with a potentially different local intrinsic dimension $d_i$. 
    
    By the law of total expectation: $\mathbb{E}_{x}[\mathcal{L}_{DSM}] = \mathbb{E}_{i} \left[ \mathbb{E}_{x \in \mathcal{M}_i} [\mathcal{L}_{DSM}(x, \sigma, \theta)] \right]$. Therefore, the global loss is lower bounded by the expected dimension of the disjoint manifolds $\mathbb{E}_{i}[d_i]$. More importantly, \Cref{cor:tightness} implies that the pointwise loss $\mathbb{E}_{\epsilon}[\mathcal{L}_{DSM}(x, \sigma, \theta)]$ direct estimator of the specific Local Intrinsic Dimension (LID) $d_i$ of the submanifold $\mathcal{M}_i$ to which $x$ belongs.
\end{remark}

\subsection{Connection with Implicit Score Matching and FLIPD}

If the LID lower bounds the denoising score matching loss, what does this imply for alternative losses such as \textit{implicit} score matching? Here, we show that the \textit{implicit} score matching loss also captures the geometric properties of the data manifold $\mathcal{M}$. The scaled \textit{implicit} score matching loss \cite{hyvarinen2005estimation} is:
\begin{equation}\label{eqn:implicit_score_matching}
    \mathbb{E}_{\noisyx \sim p(\noisyx)} \left[ \mathcal{L}_\text{ISM}(\noisyx,\, \sigma,\, \theta) \right] := \mathbb{E}_{\noisyx \sim p(\noisyx)}\left[ \sigma^2 \nabla \cdot s_\theta(\noisyx) + \frac{\sigma^2}{2}\| s_\theta(\noisyx) \|^2 \right],
\end{equation}
where $s_\theta(\vx) = -\sigma^{-1}\,\beps_\theta(\vx)$, $\sigma \in \mathbb{R}^+$, and $p(\noisyx) := (p_\rvx * \mathcal{N}({\bf 0},\, \sigma^2{\bf I}))(\noisyx)$.

\begin{theorem}[Implicit Score Matching Loss Lower Bound]\label{thm:ism_bound}
    Under the same assumptions as \Cref{thm:dsm_bound}, the implicit score matching loss is lower bounded by (proof in \Cref{sec:ism_bound_proof}):
    \begin{equation}
        \mathbb{E}_{\noisyx}[\mathcal{L}_{ISM}(\noisyx, \sigma, \theta)] \geq -(n-d) \left( \frac{\sigma^2}{\hat{\sigma}^2 + \sigma^2} \right).
    \end{equation}
\end{theorem}
The implicit score matching loss is lower bounded by a scaled negative normal dimension of $\mathcal{M}$.

We observe that FLIPD \cite{kamkari2024geometric}, the current SOTA in parametric LID estimation, is remarkably similar to the implicit score matching loss. In fact, FLIPD at a point $\vx$ is:
\begin{equation}\label{eqn:flipd}
    \text{FLIPD}(\vx,\, \sigma,\, \theta) := \mathcal{L}_\text{ISM}(\vx,\, \sigma,\, \theta) + \frac{\sigma^2}{2} \left\| s_\theta(\vx) \right\|^2 + n.
\end{equation}

FLIPD evaluates the objective on noiseless data, where the squared norm of the score is typically negligible. Substituting \Cref{thm:ism_bound} into the FLIPD formulation, dropping the non-negative score norm term, and applying the identity $n = (n-d) + d$, we yield:
\begin{equation}\label{eqn:flipd_bound}
    \mathbb{E}_{\noisyx} \left[ \text{FLIPD}(\noisyx,\, \sigma,\, \theta) \right] \geq n - (n-d) \left( \frac{\sigma^2}{\hat{\sigma}^2 + \sigma^2} \right) = d + (n-d) \left( \frac{\hat{\sigma}^2}{\hat{\sigma}^2 + \sigma^2} \right).
\end{equation}
Expected FLIPD is also lower bounded by the LID, and this bound mirrors that of the denoising score matching loss. We note the bounds in \Cref{thm:ism_bound,eqn:flipd_bound} are not tight due to dropping the score norm term, and the bound is computed with \textit{noisy} data, not \textit{noiseless} data (clean data are used in standard FLIPD). However, we believe \Cref{eqn:flipd_bound} provides useful geometric intuition on how FLIPD quantifies dimension: by measuring the normal space curvature where the score is negligible.

\subsection{Connection with the Normal Bundle Estimator}

Stanczuk et al.~\cite{stanczuk2024diffusion} propose the normal bundle (NB) LID estimator which counts the number of non-negligible singular values from a $(m, n)$ matrix $A$ of noise predictions $\beps_\theta(\noisyx)$ around a point $\vx$. This is equivalent to counting the number of non-negligible eigenvalues of the $(n,n)$ Gram matrix $C:=A^T A$, in which case the eigenvalues of $C$ are the square of the singular values of $A$.

Let us consider a variant of the NB estimator which operates on a $(m,n)$ matrix $B$ comprised of \textit{error} vectors $\left(\beps - \beps_\theta(\noisyx)\right)$. We call this the error bundle (EB) method. Let $C':= B^T B / m$ be the Gram matrix scaled by $m^{-1}$. Then the number of non-negligible eigenvalues of $C'$ should be upper bounded by $d$. Moreover, the sum of the eigenvalues of $C'$ is equivalent to the denoising score matching loss at $\vx$: $\text{trace}(C') = \mathcal{L}_\text{DSM}(\vx,\,\sigma,\, \theta) \approx d$.

\Cref{fig:eb_vs_dl} depicts this relationship for a $128$ dimensional manifold in a $256$ dimensional ambient space. The denoising loss is equal to the area under each of the spectra of $C'$ calculated from increasing numbers of samples $m$. The denoising loss predictions (vertical dashed lines) are accurate at small sample sizes (e.g., $m=8$), whereas the EB method needs at least as many samples as the LID to get a good estimate, since it counts non-negligible eigenvalues (sorted on x-axis).

\section{Experiments}

\begin{table}
  \caption{LID Estimate Mean Absolute Error (MAE) on Benchmark Manifolds}
  \label{tab:lid_benchmark}
  \centering
  \begin{tabular}{lccccccc}
    \toprule
    Parametric & \multicolumn{3}{c}{Denoising Loss (DiT)} & \multicolumn{3}{c}{FLIPD (DiT)} \\
    \midrule
     Manifold & $\sigma\seq0.01$ & $\sigma\seq0.02$ & $\sigma\seq0.05$ & $\sigma\seq0.01$ & $\sigma\seq0.02$ & $\sigma\seq0.05$ \\
    \midrule
    $d\seq16\ n\seq64$ HyperSphere & 2.00 & 2.16 & 2.58 & 1.93 & 1.59 & 0.70  \\
    $d\seq16\ n\seq64$ HyperBall & 2.57 & 2.59 & 2.71 & 0.27 & 0.51 & 3.69  \\
    $d\seq128\ n\seq256$ HyperTP & 0.47 & 1.63 & 4.48 & 1.93 & 1.30 & 1.07  \\
    $d\seq32\ n\seq128$ CliffordTorus & 17.47 & 8.47 & 4.18 & 20.15 & 11.16 & 3.30 \\
    $d\seq32\ n\seq128$ Nonlinear & 12.47 & 7.54 & 2.15 & 26.54 & 20.44 & 12.87 \\
    \midrule
    Average & 7.00 & 4.48 & \textbf{3.22} & 10.16 & 7.00 & 4.32 \\
    \toprule
    Parametric & \multicolumn{3}{c}{Denoising Loss (MLP)} & \multicolumn{3}{c}{FLIPD (MLP)} \\
    \midrule
     Manifold & $\sigma\seq0.01$ & $\sigma\seq0.02$ & $\sigma\seq0.05$ & $\sigma\seq0.01$ & $\sigma\seq0.02$ & $\sigma\seq0.05$ \\
    \midrule
    $d\seq16\ n\seq64$ HyperSphere & 2.06 & 2.07 & 2.09 & 0.67 & 1.20 & 0.95 \\
    $d\seq16\ n\seq64$ HyperBall & 2.66 & 2.66 & 2.69 & 0.35 & 0.07 & 0.20 \\
    $d\seq128\ n\seq256$ HyperTP & 0.74 & 0.55 & 0.42 & 1.27 & 7.65 & 3.02 \\
    $d\seq32\ n\seq128$ CliffordTorus & 28.90 & 28.87 & 28.76 & 32.02 & 33.03 & 30.50 \\
    $d\seq32\ n\seq128$ Nonlinear & 10.67 & 6.95 & 3.44 & 19.69 & 14.16 & 7.32 \\
    \midrule
    Average & 9.01 & 8.22 & 7.48 & 10.80 & 11.22 & 8.40 \\
    \toprule
    Non-Parametric & \multicolumn{2}{c}{MLE} & \multicolumn{2}{c}{TwoNN} & \multicolumn{2}{c}{ESS} \\
    \midrule
    Manifold & $k\seq50$ & $k\seq100$ & $k\seq50$ & $k\seq100$ & $k\seq50$ & $k\seq100$ \\
    \midrule
    $d\seq16\ n\seq64$ HyperSphere & 3.18 & 3.94 & 3.99 & 3.53 & 0.49 & 0.32 \\
    $d\seq16\ n\seq64$ HyperBall & 3.55 & 4.44 &  4.28 & 3.65 & 0.71 & 0.61  \\
    $d\seq128\ n\seq256$ HyperTP & 79.02 & 84.89 & 83.0 & 78.24 & 7.05 & 2.47 \\
    $d\seq32\ n\seq128$ CliffordTorus & 3.46 & 3.24 & 4.54 & 3.11 & 29.67 & 30.85 \\
    $d\seq32\ n\seq128$ Nonlinear & 11.68 & 13.55 & 12.74 & 11.86 & 1.30 & 1.37 \\
    \midrule
    Average & 20.18 & 22.01 & 21.71 & 20.08 & 7.84 & 7.12 \\
    \bottomrule
  \end{tabular}
\end{table}

We conduct a series of LID estimation experiments using manifolds of known local intrinsic dimension from the \texttt{scikit-dimension} package \cite{bac2021scikit}. We use the mean absolute error (MAE) across 2000 data points of true LID versus estimated LID as our main accuracy metric. We employ MLE \cite{levina2004maximum}, TwoNN \cite{facco2017estimating}, and ESS \cite{johnsson2014low} (from \texttt{scikit-dimension}) with $k=50$ and $k=100$ nearest neighbors. We train a diffusion transformer (DiT) \cite{peebles2023scalable} architecture with a patch size of 4, hidden dimension of 128, 16 attention heads, and 8 layers on each of the manifolds using a flow matching objective \cite{lipman2022flow}. We also train an MLP with skip connections, akin to \cite{kamkari2024geometric}. The model is trained on the data $2000$ points for $50000$ batches of size $100$ using a cosine annealed learning rate schedule. We convert model output (flow predictions) to noise predictions via the parameterization presented by Esser et al.~\cite{esser2024scaling}. We use the same trained model to provide LID estimates using the denoising loss and using FLIPD at $\sigma=0.01$, $\sigma=0.02$, and $\sigma=0.05$. Note that prior to estimating LID with either the denoising loss or FLIPD, the data must be scaled according to the schedule the flow matching model was trained on (scaled by $(1 - \sigma)$ in our experiments). We employ $8$ Gaussian noise samples for the denoising loss method, and we employ $8$ Rademacher samples to compute the divergence estimate for FLIPD. All experiments are implemented in PyTorch \cite{paszke2019pytorch} and run on a single NVIDIA H100 80GB GPU. For the Clifford torus, we randomly permute the dimensions of the data such that the patch-based DiT architecture must use attention to learn the manifold's structure. We center all manifold data and scale it with $\sigma_A^{-1}$, where $\sigma_A:=\max_{i} \sigma_i$ and $\sigma_i$ is the standard deviation of manifold data feature $i$.

\Cref{tab:lid_benchmark} depicts the results of the LID benchmark experiment. The non-parametric estimators perform well (MAE $<5$) on low-dimensional manifolds such as the 16-HyperSphere and 16-HyperBall, however their performance drops on highly curved, high-dimensional manifolds such as the 128-dimensional HyperTwinPeaks, 32-dimensional Clifford torus, and the 32-dimensional `Nonlinear' manifold. Of the non-parametric methods, ESS performed best with average MAE of $7.84$ and $7.12$ for $k=50$ and $k=100$, respectively.

In all combinations of architecture (MLP or DiT) and $\sigma$, the denoising loss method outperforms FLIPD in terms of average MAE. The difference in MAE between the denoising loss and FLIPD is small in most cases, but the differences can be significant on highly-curved manifolds such as `Nonlinear'. We hypothesize that this could be due to the impact of manifold curvature on the divergence term in FLIPD. Overall, the parametric methods leveraging the DiT architecture outperformed the non-parametric methods, with the denoising loss achieving an average MAE of \textbf{3.22} and FLIPD achieving an average MAE of $4.32$ for the hyperparameter choice $\sigma=0.05$.

\begin{figure*}
    \centering
    \begin{subfigure}[t]{0.49\textwidth}
        \includegraphics[width=0.99\textwidth,trim=15 0 40 20,clip]{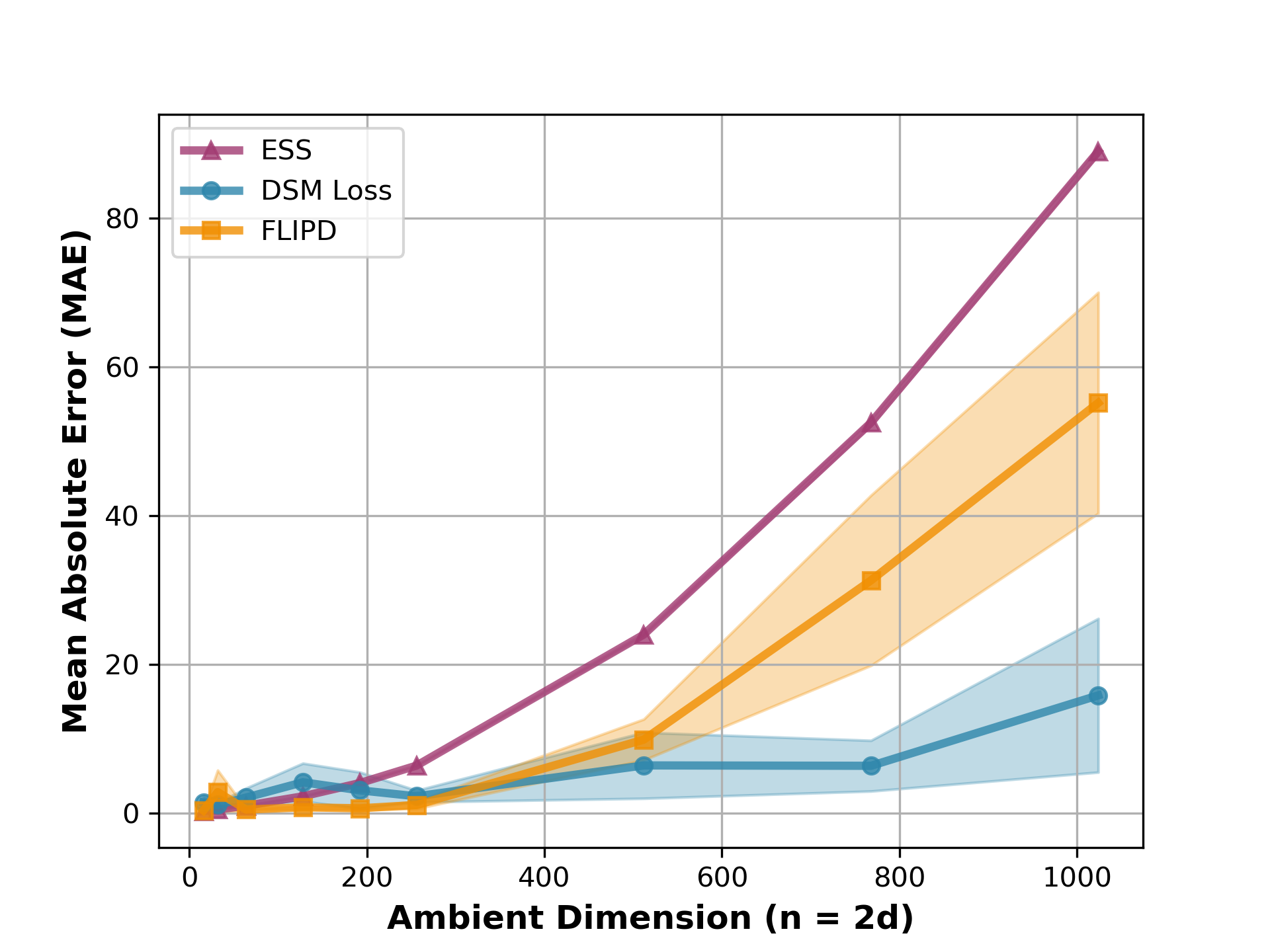}
        \caption{LID MAE ($\downarrow$) as ambient and intrinsic dimension increase.}
    \end{subfigure}
    \hfill
    \begin{subfigure}[t]{0.49\textwidth}
        \includegraphics[width=0.99\textwidth,trim=15 0 40 20,clip]{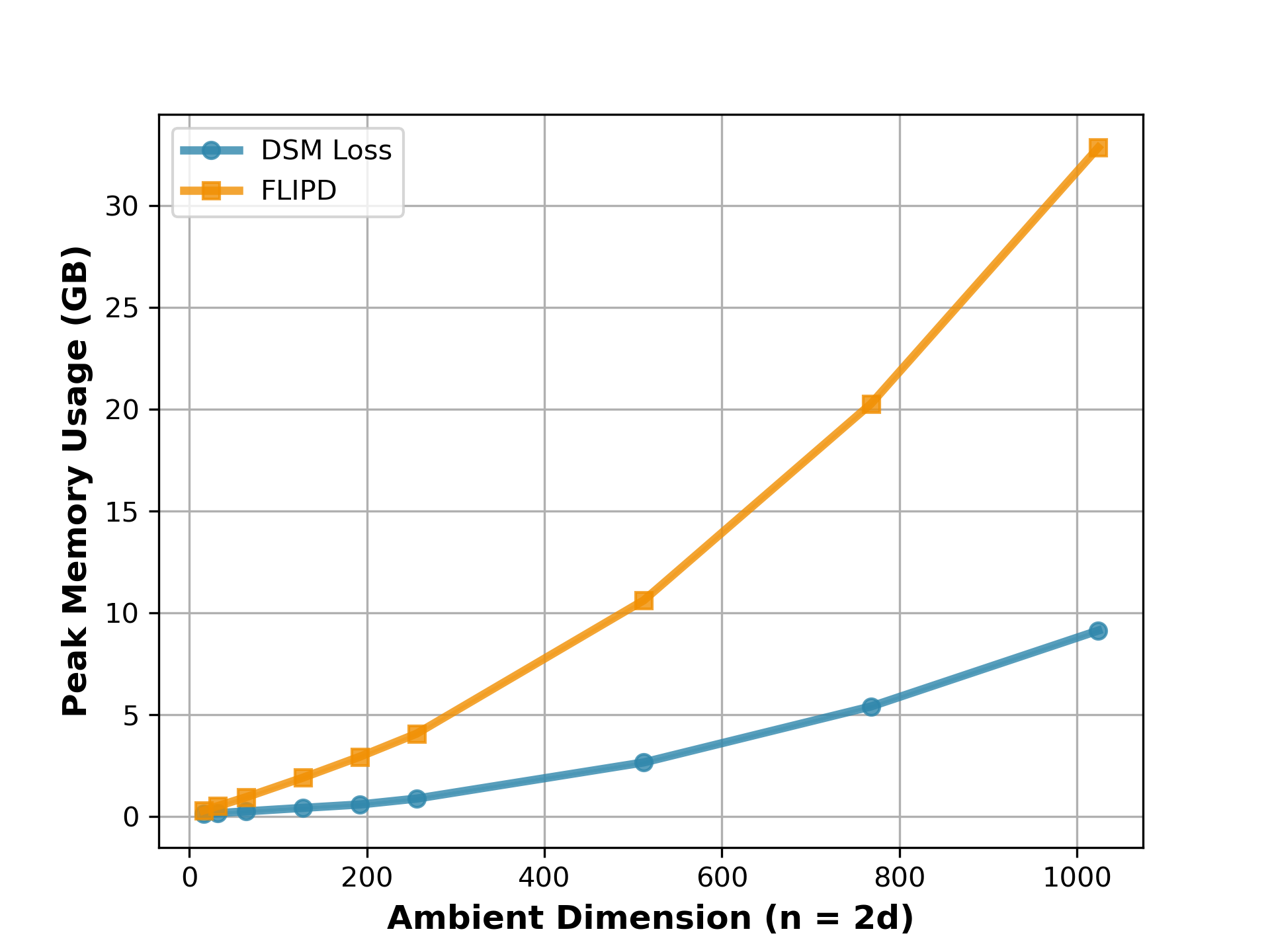}
        \caption{Peak GPU memory usage ($\downarrow$) as ambient and intrinsic dimension increase.}
    \end{subfigure}
    \caption{LID estimator accuracy and peak memory usage as ambient and intrinsic dimension increase on the Hypersphere manifold (5 trials). ESS uses $k=50$, and DSM Loss and FLIPD use $\sigma=0.1$.}\label{fig:scaling_experiment}
\end{figure*}

In \Cref{fig:scaling_experiment} we compare the MAE and peak GPU memory usage of FLIPD and the DSM loss for a sequence of hypersphere manifolds of increasing ambient dimension and LID. We employ the DiT architecture in this experiment and use 5 independent trials. For each manifold, the ambient dimension is twice the true LID of the manifold. On the left, both the DSM loss ($\sigma=0.1$) and FLIPD ($\sigma=0.1$) maintain low LID MAE until ($n=512,\,d=256)$, but only DSM loss maintains low MAE at ($n=1024,\,d=512)$. ESS ($k=50$), a non-parametric method, has low MAE on small manifolds, but it fails to scale to larger manifolds (MAE $\approx 90$ when $n=1024$ and $d=512$). On the right, we compare peak GPU memory usage for simultaneous LID computation on all $2000$ data samples for the two parametric methods. As the LID and ambient dimension of the space increases, the peak GPU memory usage for the parametric methods increases due to the use of higher dimensional data and larger DiT models. The memory usage of FLIPD increases rapidly due to its reliance on gradient computation. The peak memory usage of the DSM loss method grows more slowly, as it does not leverage gradient computation.

\subsection*{Stable Diffusion 3.5 Experiments}

Next, we implement the denoising loss method and FLIPD for LID estimation with the rectified flow transformer Stable Diffusion 3.5 medium (SD-3.5) \cite{esser2024scaling} implemented in the \texttt{diffusers} library \cite{von-platen-etal-2022-diffusers}. We sample $500$ $256 \times 256$ images of ``a photo of a cat'' from SD-3.5 with 28 sampling steps and a guidance level of $3.5$. We use the null prompt ``'' and the noise parameterization from Esser et al.~\cite{esser2024scaling} for LID estimates.

\begin{figure}[ht]
    \begin{subfigure}[t]{0.49\textwidth}
        \centering
        \includegraphics[width=0.95\textwidth]{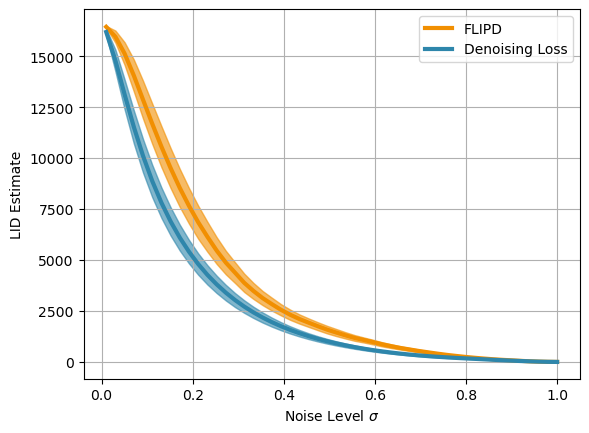}
        \caption{$\mathcal{L}_\text{DSM}$ and FLIPD LID estimates on 100 $256\times256$ images using SD-3.5.}\label{fig:lid_with_scale}
    \end{subfigure}
    \hfill
    \begin{subfigure}[t]{0.49\textwidth}
        \centering
        \includegraphics[width=0.95\textwidth]{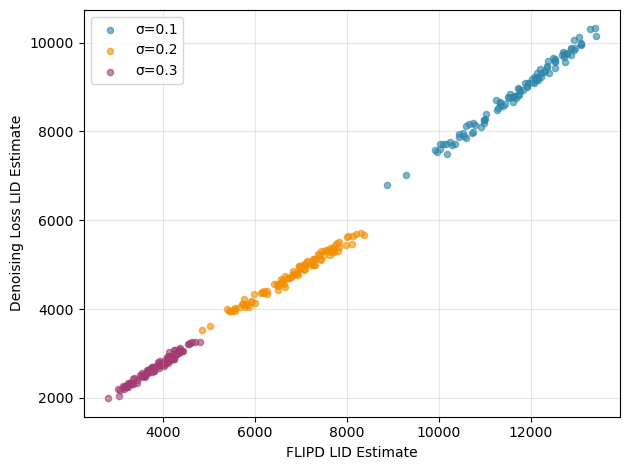}
        \caption{$\mathcal{L}_\text{DSM}$ and FLIPD LID estimate scatter plot using SD-3.5. The LID estimates are highly correlated.}\label{fig:lid_scatter}
    \end{subfigure}
    \caption{FLIPD and $\mathcal{L}_\text{DSM}$ LID estimates for 100 $256\times256$ images using SD3.5-medium. The LID estimates are highly correlated, with $\mathcal{L}_\text{DSM}$ providing lower estimates on average.}
\end{figure}

\Cref{fig:lid_with_scale} depicts the distributions of LID estimates for $100$ of the images at varying noise scales (flow matching time). Note that the LID is estimated in latent space and that the latents are scaled by $(1 - \sigma)$ for each noise level. On average, FLIPD estimates are higher than denoising loss estimates at each noise scale. At low noise scales, the data appear relatively high dimensional and occupy most of the $16384$ dimensions of the perceptually compressed latent space. Note that this is still a fraction ($< 8.4\%$) of the ambient dimension of the images. At high noise scales, the scaled data appear as a $0$-dimensional point. \Cref{fig:lid_scatter} depicts a scatter plot of LID estimates with FLIP versus LID estimates with denoising loss at various noise scales. The FLIPD and denoising loss estimates are highly correlated, and the line of best fit for each noise scale has a slope $<1$ due to FLIPD's higher-on-average LID estimates.

\begin{figure}[b]
    \begin{subfigure}[t]{0.49\textwidth}
        \centering
        \includegraphics[width=0.95\textwidth]{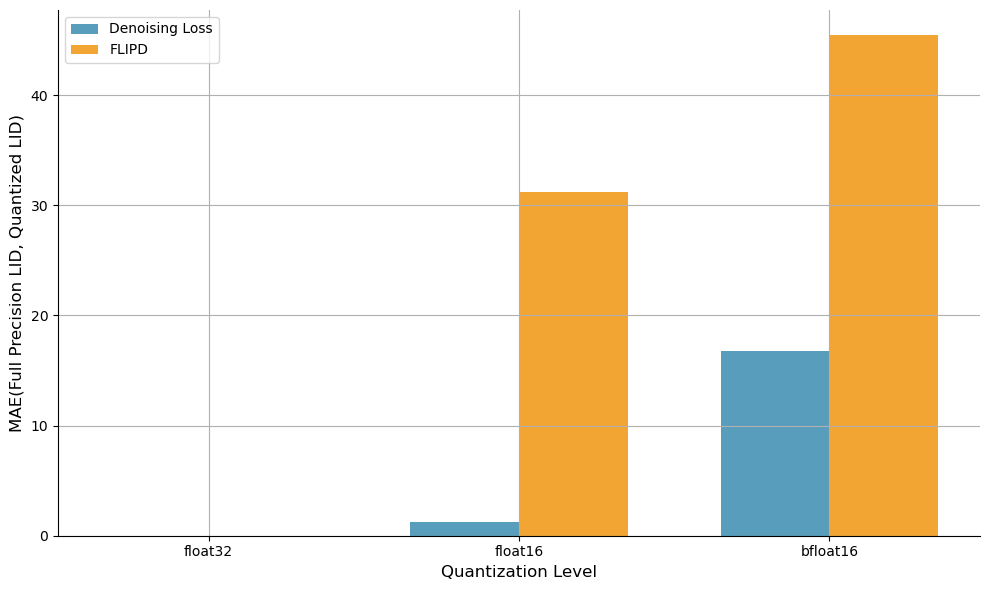}
        \caption{MAE $(\downarrow)$ of full precision LID estimates versus quantized model LID estimates.}\label{fig:sd-3.5-quant-lid}
    \end{subfigure}
    \hfill
    \begin{subfigure}[t]{0.49\textwidth}
        \centering
        \includegraphics[width=0.95\textwidth]{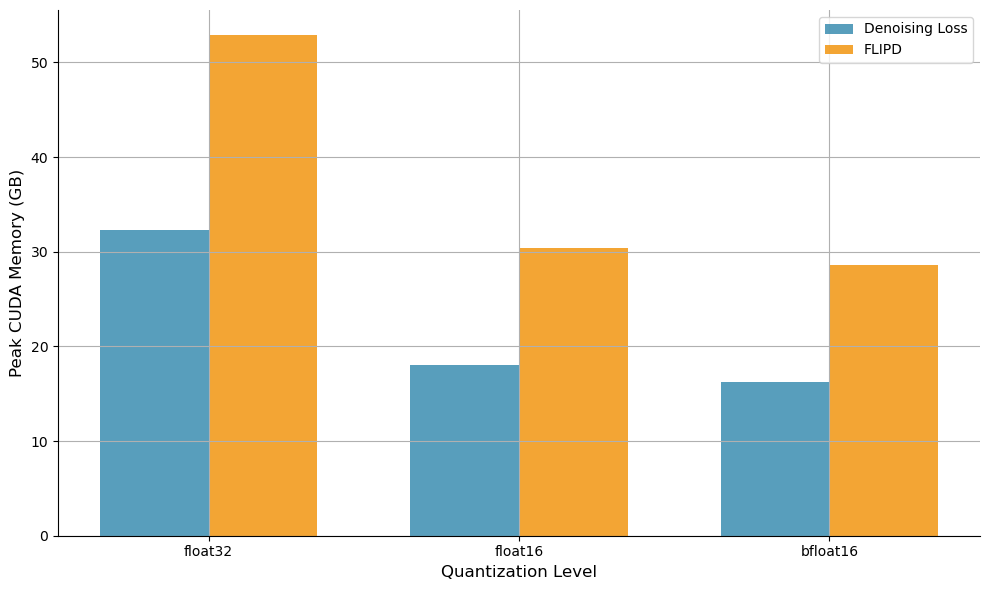}
        \caption{Average peak CUDA memory usage $(\downarrow)$ for batches of 10 $256\times256$ images.}\label{fig:sd-3.5-quant-cuda}
    \end{subfigure}
    \caption{Comparison of LID prediction disagreement (MAE) and peak GPU memory usage (GB) of SD-3.5-medium at various quantization levels.}
\end{figure}

Next, we quantize SD-3.5 (float16 and bfloat16) and record change in LID estimates (measured as MAE from float32) and peak GPU memory usage (on a batch of $10$ images) for the two parametric methods. We average results across all $500$ images in this case. In \Cref{fig:sd-3.5-quant-lid} we observe that the change in LID estimates for the denoising loss is lower than the change in FLIPD estimates after quantization. We hypothesize that this is due to accumulated error in the gradient computation that was introduced by quantization. Compared with bfloat16, the float16 quantization leads to lower MAE from float32, suggesting that LID estimation with SD-3.5 benefits more from higher precision than higher dynamic range. In \Cref{fig:sd-3.5-quant-cuda} we compare the peak GPU memory usage of FLIPD and the denoising loss on batches of $10$ images. In each case, the peak GPU memory consumption of the denoising loss is roughly $60\%$ of the peak memory consumption of FLIPD. Note that a significant portion of the memory consumption can be attributed to storing SD-3.5 on the GPU, and larger batch sizes could lead to larger differences in peak memory consumption.

\section{Discussion}

\paragraph{The Constant $C_\text{DSM}$} In this work, we show that the denoising score matching loss matches the LID of $\mathcal{M}$ in its minimum. Furthermore, recall that the minimum of $\esm$ is $0$ and that denoising model training involves amortizing the point-wise denoising loss over an entire dataset. Hence, the minimum of the average loss across a dataset is the average LID across the dataset, implying that the denoising constant $C_\text{DSM}$ from \Cref{eqn:score_matching} is the (negative) average LID of the data.

\paragraph{Interpretation of Multi-scale Training and Likelihood Computation} Diffusion and flow-based generative models are often trained on many data and noise scales which bridge the data distribution to a normal distribution \cite{song2020score,ho2020denoising,lipman2022flow,esser2024scaling}. At each data and noise scale, one can view denoising score matching training as identifying the average LID for that particular scaled and noised manifold - see figure (\Cref{fig:lid_with_scale}) for visualization. Additionally, likelihood computation for both diffusion and flow-based models is based on integration of ODEs which are expressions of the negative divergence of the learned score function through the data and noise scales. Since $\ismglobal \geq -(n-d)$ from \Cref{thm:ism_bound}, we hypothesize that one may associate higher likelihood attribution with a higher learned normal dimension (i.e., higher level of learned \textit{structure}) at each point in the ODE solution.

\paragraph{Limitations} Our experiments only use 1 H100 80GB GPU and do not leverage distributed computation to support larger batch sizes. We do not quantize models beyond half precision. We do not include ``knee search'' on LID curves \cite{kamkari2024geometric} and merely report average statistics for a few hyperparameters.

\paragraph{Societal Impacts} Our research is largely theoretical and has positive value in increasing our understanding of denoising models. We believe it does not pose any plausible negative societal impacts. The training of denoising models used in the experiments can be computationally expensive, so it caused increased electricity consumption and potential emissions.

\section{Conclusion}

We show that the denoising score matching loss is lower bounded by the local intrinsic dimension (LID) of the data manifold, motivating its use as a LID estimator. We show that the current leading LID estimator, FLIPD, is highly related to the implicit score matching loss, an objective which is equivalent to denoising score matching. Our experiments using a manifold benchmark indicate that the denoising loss is the most accurate LID estimator, and that it uses significantly less peak memory than FLIPD as the ambient dimension (problem size) increases. Lastly, our experiments with Stable Diffusion 3.5 show that the denoising loss requires less memory than FLIPD and exhibits less LID estimate degradation under quantization.




{
    \small
    \bibliographystyle{plain}
    \bibliography{references}
}


\newpage

\appendix

\section{Proofs}

\begin{definition}[Geometric and Statistical Variables]
The following variables and spaces are used in our proofs.
\begin{itemize}
    \item $n$: The ambient dimension of the data space $\mathbb{R}^n$.
    \item $d$: The intrinsic dimension of the data manifold.
    \item $\mathcal{M}$: The $d$-dimensional data manifold embedded in $\mathbb{R}^n$.
    \item $x$: A random variable drawn from the data distribution on $\mathcal{M}$.
    \item $\kappa$: The maximum principal curvature of $\mathcal{M}$ at point $x$.
    \item $\hat{\sigma}$: The scale of pre-smoothing (Gaussian noise) applied to the data, representing the non-zero "thickness" of the manifold in its normal dimensions.
    \item $\sigma$: The scale of the Gaussian noise added during the score matching process.
    \item $\epsilon$: A standard Gaussian noise vector, $\epsilon \sim \mathcal{N}(0, I_n)$.
    \item $\tilde{x}$: The observed noisy sample, defined as $\tilde{x} = x + \sigma\epsilon$.
    \item $s_\theta(\tilde{x})$: The score function parameterized by $\theta$.
    \item $\epsilon_\theta(\tilde{x})$: The noise prediction network, parameterized as $\epsilon_\theta(\tilde{x}) = -\sigma s_\theta(\tilde{x})$.
    \item $T_x\mathcal{M}$: The $d$-dimensional tangent space of $\mathcal{M}$ at $x$, spanned by orthonormal basis vectors $u_1, \dots, u_d$.
    \item $N_x\mathcal{M}$: The $(n-d)$-dimensional normal space of $\mathcal{M}$ at $x$, spanned by orthonormal basis vectors $v_1, \dots, v_{n-d}$.
\end{itemize}
\end{definition}

\subsection{Proof of \Cref{thm:dsm_bound}: Denoising Score Matching Loss Bound}\label{sec:dsm_bound_proof}

\begin{proof}
The expected scaled denoising score matching loss $\mathcal{L}_{DSM}$ tasks the model with predicting the unit-variance noise $\epsilon$:
\begin{equation}
    \mathbb{E}_{x}[\mathcal{L}_{DSM}] = \mathbb{E}_{x, \epsilon} \left[ \|\epsilon - \epsilon_\theta(\tilde{x})\|^2 \right]
\end{equation}

We decompose this error into the orthogonal components of the tangent space $T_x\mathcal{M}$ and the normal space $N_x\mathcal{M}$:
\begin{equation}
    \mathbb{E}_{x}[\mathcal{L}_{DSM}] = \mathbb{E}_{x, \epsilon} \left[ \sum_{i=1}^d \|\epsilon_i - \epsilon_\theta(\tilde{x})_i\|^2 + \sum_{j=1}^{n-d} \|\epsilon_j - \epsilon_\theta(\tilde{x})_j\|^2 \right]
\end{equation}

\textbf{1. Tangent Space Analysis:}\\
Because $\sigma \ll 1/\kappa$, the manifold is locally flat over the mass of the added noise $\sigma\epsilon$. Along the $d$ tangent dimensions $u_i$, the data distribution $p(x)$ acts as a locally uniform prior. Consequently, the observation $\tilde{x}_i = x_i + \sigma\epsilon_i$ provides no information about the specific noise realization $\epsilon_i$. This implies that the conditional distribution of the noise given the observation is identical to its marginal distribution: $p(\epsilon_i | \tilde{x}_i) = p(\epsilon_i) = \mathcal{N}(0, 1)$.

The minimum Mean Squared Error (MSE) of any estimator of a Gaussian random variable is lower-bounded by the entropy power of the target's conditional distribution. Since $h(\epsilon_i | \tilde{x}_i) = h(\epsilon_i) = \frac{1}{2}\ln(2\pi e)$, the MSE is strictly bounded by the entropy power of a standard Gaussian:
\begin{equation}
    \text{MSE}_i \ge \frac{1}{2\pi e} e^{2h(\epsilon_i | \tilde{x}_i)} = \frac{1}{2\pi e} e^{2 \left( \frac{1}{2}\ln(2\pi e) \right)} = 1
\end{equation}

Summing this irreducible error across all $d$ tangent dimensions yields the total lower bound for the tangent space:
\begin{equation}
    \mathbb{E} \left[ \sum_{i=1}^d \|\epsilon_i - \epsilon_\theta(\tilde{x})_i\|^2 \right] \ge d
\end{equation}

\textbf{2. Normal Space Analysis:}\\
Along the $n-d$ normal dimensions $v_j$, the data is pre-smoothed with thickness $\hat{\sigma}$, meaning $x_j \sim \mathcal{N}(0, \hat{\sigma}^2)$. The added noise component is $\sigma\epsilon_j$, resulting in the observation $\tilde{x}_j = x_j + \sigma\epsilon_j$, where $\tilde{x}_j \sim \mathcal{N}(0, \hat{\sigma}^2 + \sigma^2)$.

The optimal estimator that minimizes the MSE for the unit-variance noise $\epsilon_j$ is the conditional expectation $\epsilon_\theta(\tilde{x})_j = \mathbb{E}[\epsilon_j | \tilde{x}_j]$. For jointly Gaussian variables, this Minimum Mean Squared Error (MMSE) is:
\begin{equation}
    \text{MSE}_j = \text{Var}(\epsilon_j) - \frac{\text{Cov}(\epsilon_j, \tilde{x}_j)^2}{\text{Var}(\tilde{x}_j)}
\end{equation}
Given $\text{Var}(\epsilon_j) = 1$, $\text{Var}(\tilde{x}_j) = \hat{\sigma}^2 + \sigma^2$, and $\text{Cov}(\epsilon_j, x_j + \sigma\epsilon_j) = \sigma$, we substitute these values:
\begin{equation}
    \text{MSE}_j = 1 - \frac{\sigma^2}{\hat{\sigma}^2 + \sigma^2} = \frac{\hat{\sigma}^2 + \sigma^2 - \sigma^2}{\hat{\sigma}^2 + \sigma^2} = \frac{\hat{\sigma}^2}{\sigma^2 + \hat{\sigma}^2}
\end{equation}

Summing the errors from the $d$ tangent dimensions and the $n-d$ normal dimensions yields the final lower bound:
\begin{equation}
    \mathbb{E}_{x}[\mathcal{L}_{DSM}(x, \sigma, \theta)] \ge d + (n-d) \left( \frac{\hat{\sigma}^2}{\sigma^2 + \hat{\sigma}^2} \right)
\end{equation}
\end{proof}

\subsection{Proof of \Cref{thm:ism_bound}: Implicit Score Matching Loss Bound}\label{sec:ism_bound_proof}

\begin{proof}
The scaled implicit score matching loss $\mathcal{L}_{ISM}$ is defined as:
\begin{equation}
    \mathbb{E}_{\tilde{x}}[\mathcal{L}_{ISM}] = \mathbb{E}_{\tilde{x}} \left[ \sigma^2 \nabla \cdot s_\theta(\tilde{x}) + \frac{\sigma^2}{2}\|s_\theta(\tilde{x})\|^2 \right]
\end{equation}

We analyze this loss at the optimal score parameters $\theta^*$, where $s_{\theta^*}(\tilde{x}) = \nabla \log p(\tilde{x})$. We decompose the calculation into the tangent and normal spaces.

\textbf{1. Tangent Space Analysis:}\\
Because $\sigma \ll 1/\kappa$, the distribution $p(\tilde{x})$ is locally uniform along the $d$ tangent dimensions. The gradient of the log-density of a uniform distribution is zero. Thus, for each tangent dimension $u_i$:
\begin{equation}
    s_{\theta^*}(\tilde{x})_i \approx 0 \quad \text{and} \quad \nabla \cdot s_{\theta^*}(\tilde{x})_i \approx 0
\end{equation}
The contribution to the expected loss from the tangent space is bounded below by $0$.

\textbf{2. Normal Space Analysis:}\\
In each normal dimension $v_j$, the distribution of the observed noisy data is a zero-mean Gaussian with variance combining the manifold thickness and the added noise: $p(\tilde{x}_j) = \mathcal{N}(0, \hat{\sigma}^2 + \sigma^2)$. The optimal score function for this Gaussian distribution is:
\begin{equation}
    s_{\theta^*}(\tilde{x})_j = \frac{\partial}{\partial \tilde{x}_j} \log p(\tilde{x}_j) = -\frac{\tilde{x}_j}{\hat{\sigma}^2 + \sigma^2}
\end{equation}
The divergence (derivative) of this score function is:
\begin{equation}
    \frac{\partial}{\partial \tilde{x}_j} s_{\theta^*}(\tilde{x})_j = -\frac{1}{\hat{\sigma}^2 + \sigma^2}
\end{equation}

We analyze the point-wise ISM loss for a single normal dimension. Since the squared score term $\frac{1}{2} s_{\theta^*}(\tilde{x})_j^2$ is strictly non-negative, we can drop it and maintain the inequality. 

\begin{equation}
    \mathbb{E}[\mathcal{L}_{ISM}]_j = \sigma^2 \mathbb{E}_{\tilde{x}_j} \left[ \nabla \cdot s_{\theta^*}(\tilde{x})_j + \frac{1}{2} s_{\theta^*}(\tilde{x})_j^2 \right] \ge \sigma^2 \mathbb{E}_{\tilde{x}_j} \left[ \nabla \cdot s_{\theta^*}(\tilde{x})_j \right]
\end{equation}

Plugging in the divergence of the optimal score function $\nabla \cdot s_{\theta^*}(\tilde{x})_j = -\frac{1}{\hat{\sigma}^2 + \sigma^2}$:
\begin{equation}
    \mathbb{E}[\mathcal{L}_{ISM}]_j \ge \sigma^2 \left( -\frac{1}{\hat{\sigma}^2 + \sigma^2} \right) = -\frac{\sigma^2}{\hat{\sigma}^2 + \sigma^2}
\end{equation}

Summing this constant contribution over the $n-d$ normal dimensions yields the final lower bound:
\begin{equation}
    \mathbb{E}_{\tilde{x}}[\mathcal{L}_{ISM}(\tilde{x}, \sigma, \theta)] \ge -(n-d) \left( \frac{\sigma^2}{\hat{\sigma}^2 + \sigma^2} \right)
\end{equation}
\end{proof}

\subsection{The FLIPD Lower Bound}

The FLIPD estimator evaluates the implicit score matching objective on noiseless data, effectively dropping the squared score norm term and adding the ambient dimension $n$. By substituting our lower bound for $\mathcal{L}_{ISM}$ into the FLIPD formulation and dropping the additional squared score norm term, we yield:

\begin{equation}
    \mathbb{E}_{\noisyx}[\text{FLIPD}(\noisyx,\sigma,\theta)] \ge n - (n-d) \left( \frac{\sigma^2}{\hat{\sigma}^2 + \sigma^2} \right)
\end{equation}

By finding a common denominator and applying the identity $n = (n-d) + d$ to the resulting numerator, we can simplify this expression:
\begin{align}
    \mathbb{E}_{\noisyx}[\text{FLIPD}(\noisyx,\sigma,\theta)] &\ge \frac{n(\hat{\sigma}^2 + \sigma^2) - n\sigma^2 + d\sigma^2}{\hat{\sigma}^2 + \sigma^2} \nonumber \\
    &= \frac{n\hat{\sigma}^2 + d\sigma^2}{\hat{\sigma}^2 + \sigma^2} \nonumber \\
    &= \frac{((n-d) + d)\hat{\sigma}^2 + d\sigma^2}{\hat{\sigma}^2 + \sigma^2} \nonumber \\
    &= \frac{(n-d)\hat{\sigma}^2 + d(\hat{\sigma}^2 + \sigma^2)}{\hat{\sigma}^2 + \sigma^2} \nonumber \\
    &= d + (n-d) \left( \frac{\hat{\sigma}^2}{\hat{\sigma}^2 + \sigma^2} \right)
\end{align}

This simplification reveals that, under these assumptions, the theoretical lower bound for the FLIPD estimator exactly mirrors the lower bound for the Denoising Score Matching (DSM) loss. 

The bound isolates the true local intrinsic dimension $d$ alongside a strictly positive "thickness error" proportional to the normal dimensions $(n-d)$. This intuitive form demonstrates that when the data is perfectly thin ($\hat{\sigma} \to 0$), the bound tightly captures $d$. However, if the score-matching noise is chosen to be much smaller than the thickness of the data ($\sigma \ll \hat{\sigma}$), the thickness error inflates the estimate, and the estimator will inevitably capture the full ambient dimension $n$ as it resolves the local noise rather than the global manifold structure.


\subsection{Proof for \Cref{cor:tightness}: Tightness at Optimality and Zero Thickness}\label{sec:tightness_proof}

 We focus on the denoising score matching objective in this section, as the tightness at optimality for $\ism$ is clear in the proof for \Cref{thm:ism_bound}.

\begin{proof}
By the equivalence of explicit, implicit, and denoising score matching objectives \cite{vincent2011connection}, the optimal parameters $\theta^*$ minimize all three losses simultaneously. The optimal model perfectly matches the true score of the noise-perturbed data distribution: $s_{\theta^*}(\tilde{x}) = \nabla \log p_\sigma(\tilde{x})$. In the noise prediction parameterization, this is equivalent to the Minimum Mean Squared Error (MMSE) estimator: $\epsilon_{\theta^*}(\tilde{x}) = \mathbb{E}[\epsilon | \tilde{x}]$.

We evaluate the expected point-wise loss $\mathbb{E}_{\epsilon}[\|\epsilon - \epsilon_{\theta^*}(\tilde{x})\|^2]$ by decomposing it into the tangent and normal spaces at $x$.

\textbf{Tangent Space ($d$ dimensions):} By the High-Resolution Assumption, the data density acts as a uniform prior along the tangent dimensions. The noisy observation $\tilde{x}_i = x_i + \sigma\epsilon_i$ provides no information about the specific noise realization $\epsilon_i$. Consequently, the MMSE estimator simply returns the prior mean of the noise: $\mathbb{E}[\epsilon_i | \tilde{x}_i] = 0$. The resulting error per dimension is the variance of the standard Gaussian noise: $\mathbb{E}_{\epsilon_i}[\|\epsilon_i - 0\|^2] = 1$. Summing over the $d$ tangent dimensions contributes exactly $d$ to the loss.

\textbf{Normal Space ($n-d$ dimensions):} Because the manifold has zero thickness ($\hat{\sigma} = 0$), the data distribution in the normal space is a Dirac delta at the manifold surface. Hence, there is no uncertainty regarding the normal component of the clean data (in local coordinates), and the observation in normal space consists entirely of the added noise: $\tilde{x}_j = \sigma\epsilon_j$. The MMSE estimator perfectly recovers this noise: $\mathbb{E}[\epsilon_j | \tilde{x}_j] = \tilde{x}_j / \sigma = \epsilon_j$. The resulting error per dimension is $\mathbb{E}_{\epsilon_j}[\|\epsilon_j - \epsilon_j\|^2] = 0$. 

Summing the irreducible error from the tangent space and the zero error from the normal space yields a total expected point-wise loss of exactly $d + 0 = d$.
\end{proof}

\section{Example: DSM Loss on a Point Manifold with Thickness}

To provide concrete intuition for the normal space error derived in \Cref{thm:dsm_bound}, we analyze the exact optimal score and resulting expected loss for a 0-dimensional ``point manifold'' with non-zero data thickness.

\subsection{Setup and Optimal Score}
Let the data manifold $\mathcal{M}$ be a 0-dimensional point located at the origin in $\mathbb{R}^n$. Because the manifold possesses a thickness scale $\hat{\sigma}$, the true data distribution $p(x)$ is an isotropic Gaussian centered at the origin:
\begin{equation}
    p(x) = \mathcal{N}(x; \mathbf{0}, \hat{\sigma}^2 \mathbf{I})
\end{equation}

Denoising Score Matching perturbs the data by adding standard Gaussian noise scaled by $\sigma$, yielding the conditional distribution $p(\tilde{x} | x) = \mathcal{N}(\tilde{x}; x, \sigma^2 \mathbf{I})$. Because both distributions are independent zero-mean Gaussians, the marginal distribution of the noisy data $p_\sigma(\tilde{x})$ is simply their convolution:
\begin{equation}
    p_\sigma(\tilde{x}) = \mathcal{N}(\tilde{x}; \mathbf{0}, (\hat{\sigma}^2 + \sigma^2)\mathbf{I})
\end{equation}

For a model with infinite capacity, the optimal learned score $s_{\theta^*}(\tilde{x})$ matches the true score of the noise-perturbed distribution \cite{vincent2011connection}:
\begin{align}
    s_{\theta^*}(\tilde{x}) &= \nabla_{\tilde{x}} \log p_\sigma(\tilde{x}) \nonumber \\
    &= \nabla_{\tilde{x}} \left( -\frac{1}{2(\hat{\sigma}^2 + \sigma^2)} \|\tilde{x}\|^2 - \frac{n}{2} \log(2\pi(\hat{\sigma}^2 + \sigma^2)) \right) \nonumber \\
    &= -\frac{\tilde{x}}{\hat{\sigma}^2 + \sigma^2}
\end{align}

This confirms that the optimal score vector points from the noisy sample $\tilde{x}$ directly back towards the manifold (the origin), damped by the combined variance of the data thickness and the matching noise.

\subsection{Point-wise and Global Expected Loss}
We now plug this optimal score into the scaled denoising score matching loss evaluated at a specific, fixed data point $x$. Using the noise parameterization $\epsilon_{\theta^*}(\tilde{x}) = -\sigma s_{\theta^*}(\tilde{x})$ and substituting $\tilde{x} = x + \sigma\epsilon$:
\begin{align}
    \mathcal{L}_{DSM}(x, \sigma, \theta^*) &= \mathbb{E}_{\epsilon} \left[ \|\epsilon + \sigma s_{\theta^*}(\tilde{x})\|^2 \right] \nonumber \\
    &= \mathbb{E}_{\epsilon} \left[ \left\| \epsilon - \frac{\sigma}{\hat{\sigma}^2 + \sigma^2}(x + \sigma\epsilon) \right\|^2 \right] \nonumber \\
    &= \mathbb{E}_{\epsilon} \left[ \left\| \left( 1 - \frac{\sigma^2}{\hat{\sigma}^2 + \sigma^2} \right)\epsilon - \frac{\sigma}{\hat{\sigma}^2 + \sigma^2}x \right\|^2 \right] \nonumber \\
    &= \mathbb{E}_{\epsilon} \left[ \left\| \left( \frac{\hat{\sigma}^2}{\hat{\sigma}^2 + \sigma^2} \right)\epsilon - \frac{\sigma}{\hat{\sigma}^2 + \sigma^2}x \right\|^2 \right]
\end{align}

Expanding the squared $L_2$ norm yields three terms. Since $\epsilon \sim \mathcal{N}(\mathbf{0}, \mathbf{I})$, the cross-term $2\epsilon^T x$ evaluates to zero in expectation, and the expected squared norm of the noise is $\mathbb{E}[\|\epsilon\|^2] = n$. The point-wise loss becomes:
\begin{equation}
    \mathcal{L}_{DSM}(x, \sigma, \theta^*) = n \left( \frac{\hat{\sigma}^2}{\hat{\sigma}^2 + \sigma^2} \right)^2 + \frac{\sigma^2}{(\hat{\sigma}^2 + \sigma^2)^2} \|x\|^2
\end{equation}

This reveals that the local point-wise loss grows quadratically with the distance from the center of the manifold. To find the global expected loss, we take the expectation over the data distribution $x \sim \mathcal{N}(\mathbf{0}, \hat{\sigma}^2\mathbf{I})$. Noting that $\mathbb{E}_{x}[\|x\|^2] = n\hat{\sigma}^2$:
\begin{align}
    \mathbb{E}_{x}[\mathcal{L}_{DSM}(x, \sigma, \theta^*)] &= n \left( \frac{\hat{\sigma}^2}{\hat{\sigma}^2 + \sigma^2} \right)^2 + \frac{\sigma^2}{(\hat{\sigma}^2 + \sigma^2)^2} (n\hat{\sigma}^2) \nonumber \\
    &= n \hat{\sigma}^2 \left( \frac{\hat{\sigma}^2 + \sigma^2}{(\hat{\sigma}^2 + \sigma^2)^2} \right) \nonumber \\
    &= n \left( \frac{\hat{\sigma}^2}{\hat{\sigma}^2 + \sigma^2} \right)
\end{align}

Because a point manifold has an intrinsic dimension $d=0$, all $n$ ambient dimensions act as normal dimensions. Thus, this exact calculation for a 0-dimensional manifold perfectly recovers the lower bound established for normal dimensions in \Cref{thm:dsm_bound}.

\section{Stable Diffusion 2 Experiments}

\begin{figure}[ht]
    \begin{subfigure}[t]{0.49\textwidth}
        \centering
        \includegraphics[width=0.95\textwidth]{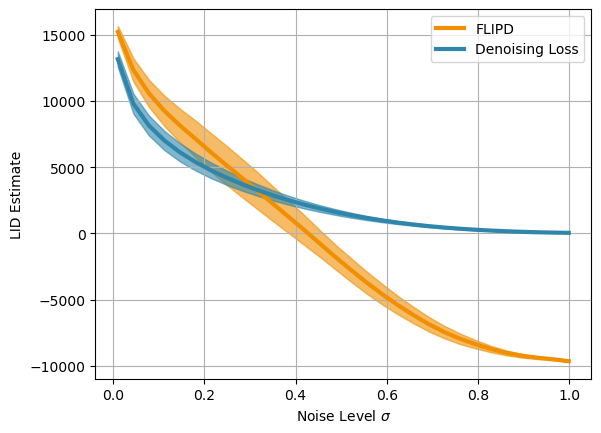}
        \caption{$\mathcal{L}_\text{DSM}$ and FLIPD LID estimates on 500 $512\times512$ images using SD2.}\label{fig:lid_with_scale_sd2}
    \end{subfigure}
    \hfill
    \begin{subfigure}[t]{0.49\textwidth}
        \centering
        \includegraphics[width=0.95\textwidth]{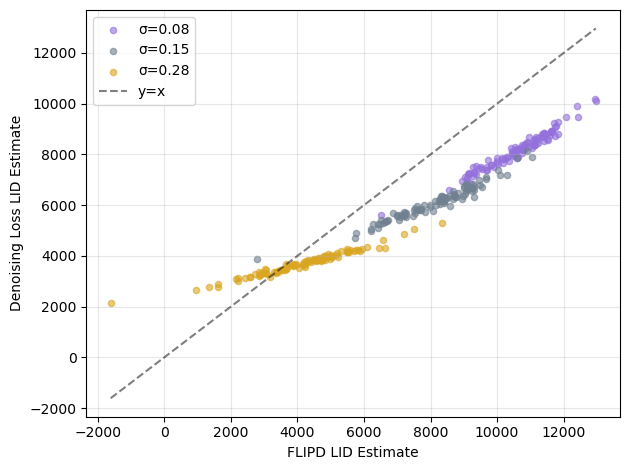}
        \caption{$\mathcal{L}_\text{DSM}$ and FLIPD LID estimate scatter plot for 100 samples using SD2. The LID estimates are highly correlated.}\label{fig:lid_scatter_sd2}
    \end{subfigure}
    \caption{FLIPD and $\mathcal{L}_\text{DSM}$ LID estimates for $512\times512$ images using Stable Diffusion 2 (SD2).  The LID estimates are highly correlated, with $\mathcal{L}_\text{DSM}$ providing lower estimates at low noise levels.}\label{fig:sd2_experiment}
\end{figure}

We include a similar experiment comparing FLIPD and the denoising loss using Stable Diffusion 2 (SD2) \cite{rombach2022high}. SD2 is a latent diffusion model which uses a U-Net \cite{ronneberger2015u} architecture. 
\Cref{fig:sd2_experiment} depicts the results of the experiment. Like with SD3.5, the LID estimates are highly correlated, and FLIPD typically assigns higher LID estimates at low noise levels. We note that the FLIPD LID estimates become invalid (negative) at higher noise levels. Note that this is possible because FLIPD is calculated on \textit{noiseless} data, and the bound \Cref{eqn:flipd_bound} only applies to $\sigma-$noisy data. We hypothesize that the negative LID estimates do not occur for FLIPD with SD3.5 due to the superior noise parameterization $\eps_\theta(\vx,\, \sigma) := (1-\sigma) v_\theta(\vx,\, \sigma) + \vx$ of flow matching models \cite{lipman2022flow,esser2024scaling}.

\end{document}